\documentclass[runningheads]{llncs}
\usepackage{graphicx}

\usepackage{tikz}
\usepackage{comment}
\usepackage{amsmath} 
\usepackage{color}
\usepackage{bm}
\usepackage{bbding}
\usepackage{graphicx}
\usepackage{amssymb}
\usepackage{multirow}
\usepackage{gensymb}
\usepackage{booktabs}
\usepackage{colortbl}
\usepackage{nicematrix}
\usepackage{url}
\usepackage[pagebackref=true,breaklinks=true,letterpaper=true,colorlinks,bookmarks=false]{hyperref}

\usepackage[accsupp]{axessibility}  

\newcolumntype{C}[1]{>{\PreserveBackslash\centering}p{#1}}

\begin{document}
\pagestyle{headings}
\mainmatter

\title{Category-Level 6D Object Pose and Size Estimation using Self-Supervised Deep Prior Deformation Networks} 


\titlerunning{Self-Supervised Deep Prior Deformation Network}

\author{Jiehong Lin\inst{1,2} \and
Zewei Wei\inst{1}\and
Changxing Ding\inst{1} \and
Kui Jia\inst{1,3}\thanks{Corresponding author}}
\authorrunning{J. Lin et al.}
\institute{South China University of Technology \and
DexForce Co. Ltd. \and
Peng Cheng Laboratory
\\
\email{\{lin.jiehong,eeweizewei\}@mail.scut.edu.cn; \{chxding,kuijia\}@scut.edu.cn}}

\maketitle

\begin{abstract}
It is difficult to precisely annotate object instances and their semantics in 3D space, and as such, synthetic data are extensively used for these tasks, e.g., category-level 6D object pose and size estimation. However, the easy annotations in synthetic domains bring the downside effect of synthetic-to-real (Sim2Real) domain gap. In this work, we aim to address this issue in the task setting of Sim2Real, unsupervised domain adaptation for category-level 6D object pose and size estimation. We propose a method that is built upon a novel \textit{Deep Prior Deformation Network}, shortened as DPDN. DPDN learns to deform features of categorical shape priors to match those of object observations, and is thus able to establish deep correspondence in the feature space for direct regression of object poses and sizes. To reduce the Sim2Real domain gap, we formulate a novel self-supervised objective upon DPDN via consistency learning; more specifically, we apply two rigid transformations to each object observation in parallel, and feed them into DPDN respectively to yield dual sets of predictions; on top of the parallel learning, an inter-consistency term is employed to keep cross consistency between dual predictions for improving the sensitivity of DPDN to pose changes, while individual intra-consistency ones are used to enforce self-adaptation within each learning itself. We train DPDN on both training sets of the synthetic CAMERA25 and real-world REAL275 datasets; our results outperform the existing methods on REAL275 test set under both the unsupervised and supervised settings. Ablation studies also verify the efficacy of our designs. Our code is released publicly at \url{https://github.com/JiehongLin/Self-DPDN}.
\end{abstract}

\section{Introduction}

The task of category-level 6D object pose and size estimation, formally introduced in \cite{NOCS}, is to estimate the rotations, translations, and sizes of unseen object instances of certain categories in cluttered RGB-D scenes. It plays a crucial role in many real-world applications, such as robotic grasping\cite{wu2020grasp,mousavian20196}, augmented reality\cite{azuma1997survey}, and autonomous driving \cite{levinson2011towards,wang2019frustum,chen2017multi,deng2022vista}.

For this task, existing methods can be roughly categorized into two groups, i.e., those based on direct regression and those based on dense correspondence learning. Methods of the former group \cite{FSNet,DualPoseNet,SS-Conv} are conceptually simple, but struggle in learning  pose-sensitive features such that direct predictions can be made in the full $SE(3)$ space; dense correspondence learning \cite{NOCS,SPD,CRNet,SGPA,UDA-COPE} makes the task easier by first regressing point-wise coordinates in the canonical space to align with points of observations, and then obtaining object poses and sizes via solving of Umeyama algorithm \cite{Umeyama}. Recent works \cite{SPD,CRNet,SGPA,UDA-COPE} of the second group exploit strong categorical priors (\textit{e.g.}, mean shapes of object categories) for improving the qualities of canonical point sets, and constantly achieve impressive results; however, their surrogate objectives for the learning of canonical coordinates are one step away from the true ones for estimating object poses and sizes, making their learning suboptimal to the end task.

The considered learning task is further challenged by the lack of real-world RGB-D data with careful object pose and size annotations in 3D space. As such, synthetic data are usually simulated and rendered whose annotations can be freely obtained on the fly \cite{NOCS,denninger2020blenderproc}. However, the easy annotations in synthetic domains bring a downside effect of synthetic-to-real (Sim2Real) domain gap; learning with synthetic data with no consideration of Sim2Real domain adaptation would inevitably result in poor generalization in the real-world domain. This naturally falls in the realm of Sim2Real, unsupervised domain adaptation (UDA) \cite{ajakan2014domain,long2015learning,qin2019pointdan,zhang2019domain,zhang2020label,zhang2020unsupervised,wang2020self6d,UDA-COPE}.
 
\begin{figure}[t]
\centering
\includegraphics[width=0.92\linewidth]{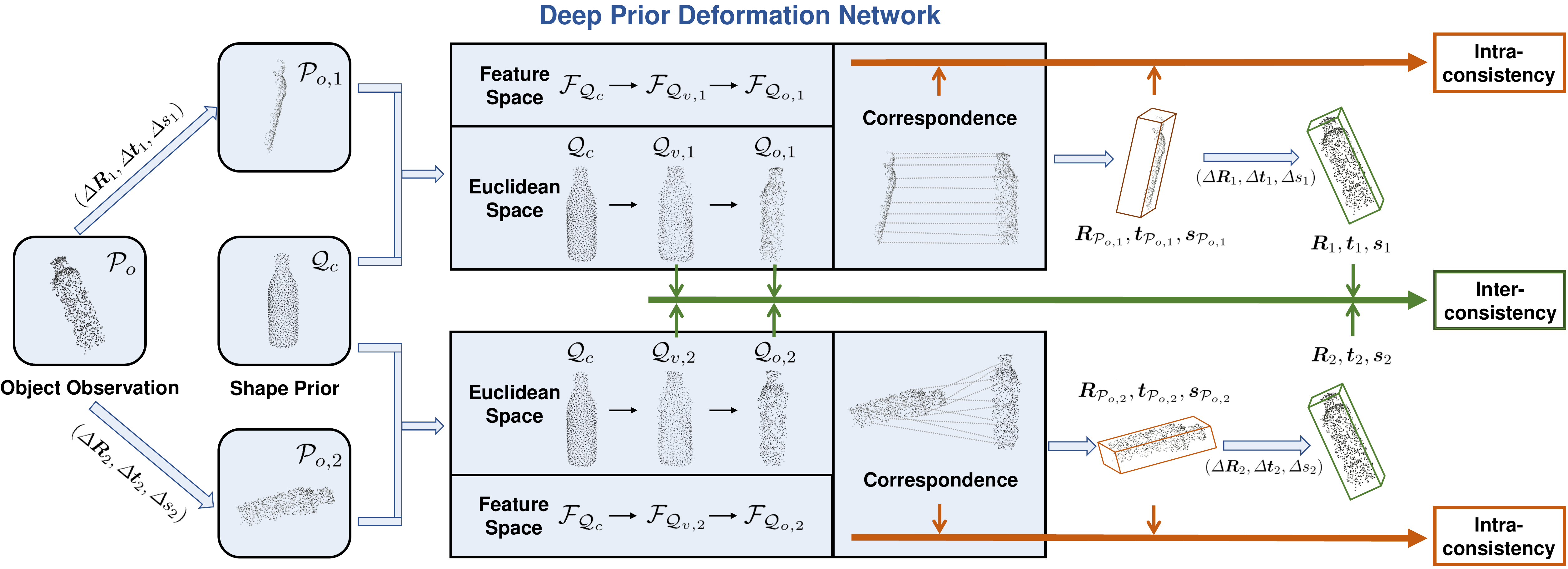}
\caption{An illustration of our proposed self-supervised Deep Prior Deformation Network (DPDN). DPDN deforms categorical shape priors in the feature space to pair with object observations, and establishes deep correspondence for direct estimates of object poses and sizes; upon DPDN, a novel self-supervised objective is designed to reduce synthetic-to-real domain gap via consistency learning. Specifically, we apply two rigid transformations to the point set $\mathcal{P}_{o}$ of an object observation, and feed them into DPDN in parallel to make dual sets of predictions; on top of the parallel learning, an inter-consistency term between dual predictions is then combined with individual intra-consistency ones within each learning to form the self-supervision. For simplicity, we omit the image input of object observation. Notations are explained in Sec. \ref{sec:method}.}
\label{fig:head}
\end{figure}

In this work, we consider the task setting of Sim2Real UDA for category-level 6D object pose and size estimation. We propose a new method of \textit{self-supervised Deep Prior Deformation Network}; Fig. \ref{fig:head} gives an illustration. Following dense correspondence learning, we first present a novel \textit{Deep Prior Deformation Network}, shortened as \textbf{DPDN}, which implements a deep version of shape prior deformation in the feature space, and is thus able to establish deep correspondence for direct regression of poses and sizes with high precision. For a cluttered RGB-D scene, we employ a 2D instance segmentation network (\textit{e.g}, Mask RCNN \cite{MasRCNN}) to segment the objects of interest out, and feed them into our proposed DPDN for pose and size estimation. As shown in Fig. \ref{fig:DPDN}, the architecture of DPDN consists of three main modules, including a Triplet Feature Extractor, a Deep Prior Deformer, and a Pose and Size Estimator. For an object observation, the Triplet Feature Extractor learns point-wise features from its image crop, point set, and categorical shape prior, respectively; then Deep Prior Deformer deforms the prior in feature space by learning a feature deformation field and a correspondence matrix, and thus builds deep correspondence from the observation to its canonical version; finally, Pose and Size Estimator is used to make reliable predictions directly from those built deep correspondence.

On top of DPDN, we formulate a self-supervised objective that combines an inter-consistency term with two intra-consistency ones for UDA. More specifically, as shown in Fig. \ref{fig:head}, we apply two rigid transformations to an input point set of object observation, and feed them into our DPDN in parallel for making dual sets of predictions. Upon the above parallel learning, the inter-consistency term enforces cross consistency between dual predictions w.r.t. two transformations for improving the sensitivity of DPDN to pose changes, and within each learning, the individual intra-consistency term is employed to enforce self-adaptation between the correspondence and the predictions.
We train DPDN on both training sets of the synthetic CAMERA25 and real-world REAL275 datasets \cite{NOCS}; our results outperform the existing methods on REAL275 test set under both unsupervised and supervised settings. We also conduct ablation studies that confirm the advantages of our designs. Our contributions can be summarized as follows:
\begin{itemize}
    \item We propose a \textit{Deep Prior Deformation Network}, termed as DPDN, for the task of category-level 6D object pose and size estimation. DPDN deforms categorical shape priors to pair with object observations in the feature space, and is thus able to establish deep correspondence for direct regression of object poses and sizes.

    \item Given that the considered task largely uses synthetic training data, we formulate a novel self-supervised objective upon DPDN to reduce the synthetic-to-real domain gap. The objective is built upon enforcing consistencies between parallel learning w.r.t. two rigid transformations, and has the effects of both improving the sensitivity of DPDN to pose changes, and making predictions more reliable.

    \item We conduct thorough ablation studies to confirm the efficacy of our designs. Notably, our method outperforms existing ones on the benchmark dataset of real-world REAL275 under both the unsupervised and supervised settings.
\end{itemize}

\section{Related Work}

\subsubsection{Fully-Supervised Methods} 

Methods of fully-supervised category-level 6D pose and size estimation could be roughly divided into two groups, \textit{i.e.}, those based on direct regression \cite{FSNet,DualPoseNet,SS-Conv} and those based on dense correspondence learning \cite{NOCS,SPD,CRNet,SGPA}. 

Direct estimates of object poses and sizes from object observations suffer from the difficulties in the learning of the full $SE(3)$ space, and thus make demands on extraction of pose-sensitive features. FS-Net \cite{FSNet} builds an orientation-aware backbone with 3D graph convolutions to encode object shapes, and makes predictions with a decoupled rotation mechanism. DualPoseNet \cite{DualPoseNet} encodes pose-sensitive features from object observations based on rotation-equivariant spherical convolutions, while two parallel pose decoders with different working mechanisms are stacked to impose complementary supervision. A recent work of SS-ConvNet \cite{SS-Conv} designs Sparse Steerable Convolutions (SS-Conv) to further explore SE(3)-equivariant feature learning, and presents a two-stage pose estimation pipeline upon SS-Convs for iterative pose refinement.

Another group of works first learn coordinates of object observations in the canonical space to establish dense correspondence, and then obtain object poses and sizes by solving Umeyama algorithm from the correspondence in 3D space. NOCS \cite{NOCS}, the first work for our focused task, is realized in this way by directly regressing canonical coordinates from RGB images. SPD \cite{SPD} then makes the learning of canonical points easier by deforming categorical shape priors, rather than directly regressing from object observations. The follow-up works also confirm the advantages of shape priors, and make efforts on the prior deformation to further improve the qualities of canonical points, \textit{e.g.}, via recurrent reconstruction for iterative refinement \cite{CRNet}, or structure-guided adaptation based on transformer \cite{SGPA}.

\subsubsection{Unsupervised Methods}

Due to the time-consuming and labor-intensive annotating of real-world data in 3D space, UDA-COPE \cite{UDA-COPE} presents a new setting of unsupervised domain adaptation for the focused task, and adapts a teacher-student scheme with bidirectional point filtering to this setting, which, however, heavily relies on the qualities of pseudo labels. In this paper, we exploit inter-/intra-consistency in the self-supervised objective to explore the data characteristics of real-world data and fit the data for the reduction of domain gap.

\section{Self-Supervised Deep Prior Deformation Network} 

\label{sec:method}

Given a cluttered RGB-D scene, the goal of \textit{Category-Level 6D Object Pose and Size Estimation} is to detect object instances of interest with compact 3D bounding boxes, each of which is represented by rotation $\bm{R}\in SO(3)$, translation $\bm{t}\in\mathbb{R}^3$, and size $\bm{s} \in \mathbb{R}^3$ \textit{w.r.t.} categorical canonical space. 

A common practice to deal with this complicated task is decoupling it into two steps, including 1) object detection/instance segmentation, and 2) object pose and size estimation. For the first step, there exist quite mature techniques to accomplish it effectively, \textit{e.g.}, employing an off-the-shelf MaskRCNN \cite{MasRCNN} to segment object instances of interest out; for the second step, however, it is still challenging to directly regress poses of unknown objects, especially for the learning in $SO(3)$ space. To settle this problem, we propose in this paper a novel \textit{Deep Prior Deformation Network} $\mathrm{\Phi}$, shortened as \textbf{DPDN}, which deforms categorical shape priors to match object observations in the feature space, and estimates object poses and sizes from the built deep correspondence directly; Fig. \ref{fig:DPDN} gives an illustration. We will detail the architecture of $\mathrm{\Phi}$ in Sec. \ref{subsec: DPDN_architecture}.

Another challenge of this task is the difficulty in precisely annotating real-world data in 3D space. Although synthetic data at scale are available for the learning of deep models, their results are often less precise than those trained with annotated real-world data, due to downside effect of large domain gap. To this end, we take a mixture of labeled synthetic data and unlabeled real-world one for training, and design a novel self-supervised objective for synthetic-to-real (Sim2Real), unsupervised domain adaptation. Specifically, given a mini batch of $B$ training instances $\{\mathcal{V}_i\}_{i=1}^B$, we solve the following optimization problem on top of $\mathrm{\Phi}$:
\begin{equation}
    \min_{\mathrm{\Phi}}  \sum_{i=1}^{B} \frac{1}{B_{1}} \alpha_{i} \mathcal{L}_{supervised}^{\mathcal{V}_i} + \frac{1}{B_2} (1-\alpha_{i}) \mathcal{L}_{self-supervised}^{\mathcal{V}_i},
\end{equation}
with $B_{1}=\sum_{i=1}^{B}\alpha_{i}$ and $B_{2}=\sum_{i=1}^{B}1-\alpha_{i}$.
$\{\alpha_{i}\}_{i=1}^{B}$ is a binary mask; $\alpha_{i}=1$ if the observation of $\mathcal{V}_i$ is fully annotated and $\alpha_{i}=0$ otherwise. In Sec. \ref{subsec: DPDN_self_supervised}, we will give a detailed illustration on the self-supervised objective $\mathcal{L}_{self-supervised}$, which learns inter-consistency and intra-consistency upon DPDN, while the illustration on the supervised objective $\mathcal{L}_{supervised}$ is included in Sec. \ref{subsec: DPDN_supervised}.

\subsection{Deep Prior Deformation Network}
\label{subsec: DPDN_architecture}

\begin{figure}[t]
\centering
\includegraphics[width=0.98\linewidth]{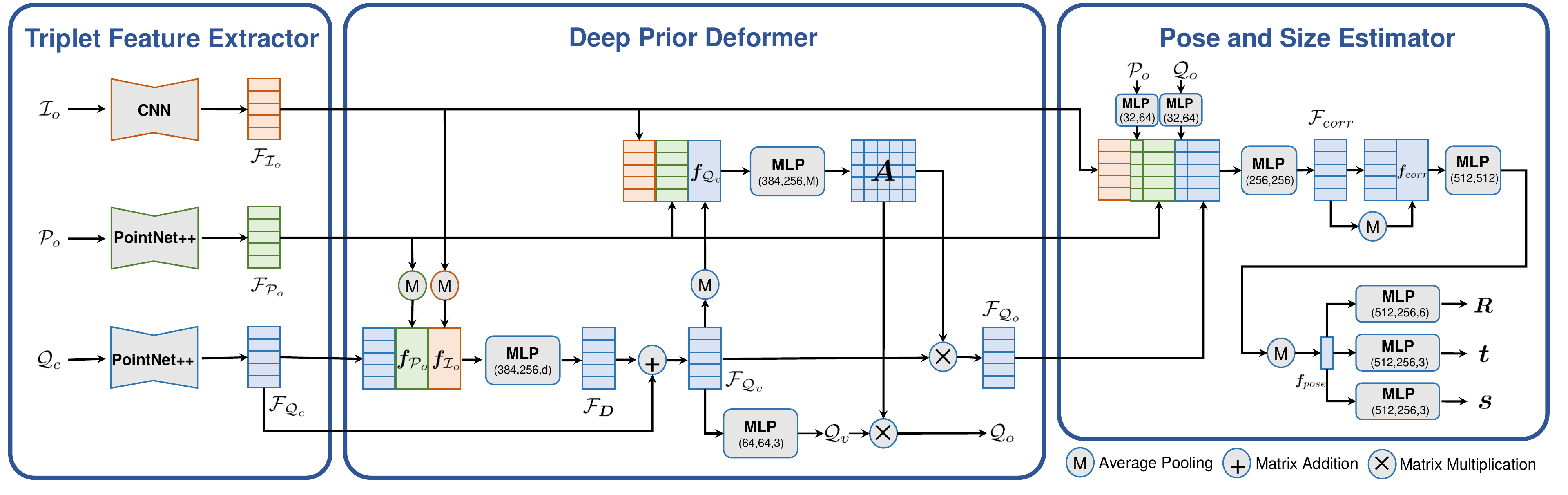}
\caption{An illustration of our Deep Prior Deformation Network (DPDN). For an object observation $\mathcal{V}$, we take its image crop $\mathcal{I}_o$, point set $\mathcal{P}_o$, and shape prior $\mathcal{Q}_c$ of the same category as inputs of a \textbf{Triplet Feature Extractor}, for the learning of their point-wise features $\mathcal{F}_{\mathcal{I}_o}$, $\mathcal{F}_{\mathcal{P}_o}$, and $\mathcal{F}_{\mathcal{Q}_c}$, respectively; then a \textbf{Deep Prior Deformer} is employed to learn a feature deformation field $\mathcal{F}_{\bm{D}}$ and a correspondence matrix $\bm{A}$ to deform $\mathcal{F}_{\mathcal{Q}_c}$, yielding a feature map $\mathcal{F}_{\mathcal{Q}_o}$ in the canonical space to pair with $\mathcal{F}_{\mathcal{P}_o}$; a \textbf{Pose and Size Estimator} makes final predictions $(\bm{R}, \bm{t}, \bm{s})$ directly from the built deep correspondence between $\mathcal{F}_{\mathcal{P}_o}$ and $\mathcal{F}_{\mathcal{Q}_o}$.}
\label{fig:DPDN}
\end{figure}

For an object instance $\mathcal{V}$ belonging to a category $c$ of interest, we represent its RGB-D observation in the scene as $(\mathcal{I}_{o}, \mathcal{P}_{o})$, where $\mathcal{I}_{o} \in \mathbb{R}^{H\times W \times 3}$ denotes the RGB segment compactly containing the instance with a spatial size of $H\times W$, and $\mathcal{P}_{o} \in \mathbb{R}^{N \times 3}$ denotes the masked point set with N object surface points. Direct regression of object pose and size from $(\mathcal{I}_{o}, \mathcal{P}_{o})$ struggles in the learning of SO(3) space without object CAD model \cite{FSNet,DualPoseNet,SS-Conv}. 

Alternatively, a recent group of works \cite{SPD,CRNet,SGPA,UDA-COPE} achieve impressive results by taking advantages of strong categorical priors to establish dense point-wise correspondence, from which object pose and size could be obtained by solving of Umeyama algorithm \cite{Umeyama}. Specifically, assuming $\mathcal{Q}_{c} \in \mathbb{R}^{M\times 3}$ is a sampled point set from the shape prior of $c$ with $M$ points, a point-wise deformation field $\bm{D} \in \mathbb{R}^{ M \times 3}$ and a correspondence matrix $\bm{A} \in \mathbb{R}^{N\times M}$ are learned from a triplet of $(\mathcal{I}_{o}, \mathcal{P}_{o}, \mathcal{Q}_{c})$. $\bm{D}$ contains point-wise deviations with respect to $\mathcal{Q}_{c}$, deforming $\mathcal{Q}_{c}$ to $\mathcal{Q}_{v} \in \mathbb{R}^{M\times 3}$, which represents a complete shape of $\mathcal{V}$ in the canonical space. $\bm{A}$ models relationships between points in $\mathcal{Q}_{v}$ and $\mathcal{P}_{o}$, serving as a sampler from $\mathcal{Q}_{v}$ to generate a partial point set $\mathcal{Q}_{o}\in \mathbb{R}^{N\times 3}$, paired with $\mathcal{P}_{o}$, as follows:
\begin{eqnarray}
    \mathcal{Q}_{o}  = \bm{A} \times \mathcal{Q}_{v} = \bm{A} \times (\mathcal{Q}_{c} + \bm{D} ).
\label{Eqn:ShallowShapeDeformation}
\end{eqnarray}
Finally, solving of Umeyama algorithm to align $\mathcal{Q}_{o}$ with $\mathcal{P}_{o}$ gives out the target pose and size.

However, surrogate objectives for the learning of $\bm{A}$ and $\bm{D}$ are a step away from the true ones for estimates of pose and size; for example, small deviations of $\bm{A}$ or $\bm{D}$ may lead to large changes in the pose space. Thereby, we present a Deep Prior Deformation Network (DPDN), which implements a deep version of (\ref{Eqn:ShallowShapeDeformation}) as follows:
\begin{equation}
    \mathcal{F} _{\mathcal{Q}_{o}} = \bm{A} \times \mathcal{F}_{\mathcal{Q}_{v}} = \bm{A} \times (\mathcal{F}_{\mathcal{Q}_{c}} + \mathcal{F}_{\bm{D}}),
\label{Eqn:DeepShapeDeformation}
\end{equation}
where $\mathcal{F}_{\mathcal{Q}_{c}}$, $\mathcal{F}_{\mathcal{Q}_{v}}$, and $\mathcal{F} _{\mathcal{Q}_{o}}$ denote point-wise features of $\mathcal{Q}_{c}$, $\mathcal{Q}_{v}$, and $\mathcal{Q}_{o}$, respectively, and $\mathcal{F}_{\bm{D}}$ is a feature deformation field \textit{w.r.t.} $\mathcal{F}_{\mathcal{Q}_{c}}$. The deep version (\ref{Eqn:DeepShapeDeformation}) deforms $\mathcal{Q}_{c}$ in the feature space, such that features of $\mathcal{Q}_{o}$ and $\mathcal{P}_{o}$ are paired to establish deep correspondence, from which object pose and size $(\bm{R}, \bm{t}, \bm{s})$ could be predicted via a subsequent network. Direct regression from deep correspondence thus alleviates the difficulties encountered by that from object observation. We note that upon the correspondence and the predictions, a self-supervised signal of intra-consistency could also be built for unlabeled data (see Sec. \ref{subsec: DPDN_self_supervised}).

As depicted in Fig. \ref{fig:DPDN}, the architecture of DPDN consists of three main modules, including \textbf{Triplet Feature Extractor}, \textbf{Deep Prior Deformer}, and \textbf{Pose and Size Estimator}. We will give detailed illustrations shortly.

\subsubsection{Triplet Feature Extractor}

Given the inputs of object observation $(\mathcal{I}_{o}, \mathcal{P}_{o})$ and categorical shape prior $\mathcal{Q}_{c}$, we firstly extract their point-wise features $(\mathcal{F}_{\mathcal{I}_{o}}, \mathcal{F}_{\mathcal{P}_{o}})\in (\mathbb{R}^{N\times d}, \mathbb{R}^{N\times d})$ and $\mathcal{F}_{\mathcal{Q}_{c}}\in \mathbb{R}^{M\times d}$, where $d$ denotes the number of feature channels. Following \cite{wang2019densefusion}, we firstly employ a PSP network \cite{PSPNet} with ResNet-18 \cite{ResNet} to learn pixel-wise appearance features of $\mathcal{I}_{o}$, and then select those corresponding to $\mathcal{P}_{o}$ out to form $\mathcal{F}_{\mathcal{I}_{o}}$. For both $\mathcal{P}_{o}$ and $\mathcal{Q}_{c}$, two networks of PointNet++ \cite{pointnet++} decorated with 4 set abstract levels are individually applied to extract their point-wise geometric features $\mathcal{F}_{\mathcal{P}_{o}}$ and $\mathcal{F}_{\mathcal{Q}_{c}}$.


\subsubsection{Deep Prior Deformer}

After obtaining $\mathcal{F}_{\mathcal{I}_{o}}$, $\mathcal{F}_{\mathcal{P}_{o}}$ and $\mathcal{F}_{\mathcal{Q}_{c}}$, the goal of Deep Prior Deformer is to learn the feature deformation field $\mathcal{F}_{\bm{D}} \in \mathbb{R}^{M \times d}$ and the correspondence matrix $\bm{A} \in \mathbb{R}^{N\times M}$ in (\ref{Eqn:DeepShapeDeformation}), and then implement (\ref{Eqn:DeepShapeDeformation}) in the feature space to establish deep correspondence.

Specifically, as shown in Fig. \ref{fig:DPDN}, we obtain global feature vectors $\bm{f}_{\mathcal{I}_{o}} \in \mathbb{R}^d$ and $\bm{f}_{\mathcal{P}_{o}} \in \mathbb{R}^d $ of $\mathcal{I}_{o}$ and $\mathcal{P}_{o}$, respectively, by averaging their point-wise features $\mathcal{F}_{\mathcal{I}_{o}}$ and $\mathcal{F}_{\mathcal{P}_{o}}$; then each point feature of $\mathcal{F}_{\mathcal{Q}_{c}}$ is fused with $\bm{f}_{\mathcal{I}_{o}}$ and $\bm{f}_{\mathcal{P}_{o}}$, and fed into a subnetwork of Multi-Layer Perceptron (MLP) to learn its deformation. Collectively, the whole feature deformation field $\mathcal{F}_{\bm{D}}$ could be learned as follows:
\begin{eqnarray}
    \begin{matrix}
        \mathcal{F}_{\bm{D}} = \texttt{MLP}([\mathcal{F}_{\mathcal{Q}_{c}}, \texttt{Tile}^{M}(\bm{f}_{\mathcal{I}_{o}}), \texttt{Tile}^{M}(\bm{f}_{\mathcal{P}_{o}})]),\\
        s.t. \quad \bm{f}_{\mathcal{I}_{o}} = \texttt{AvgPool}(\mathcal{F}_{\mathcal{I}_{o}}), \quad  \bm{f}_{\mathcal{P}_{o}} = \texttt{AvgPool}(\mathcal{F}_{\mathcal{P}_{o}}),
    \end{matrix}
\end{eqnarray}
where $[\cdot, \cdot]$ denotes concatenation along feature dimension, $\texttt{MLP}(\cdot)$ denotes a trainable subnetwork of MLP, $\texttt{AvgPool}(\cdot)$ denotes an average-pooling operation over surface points, and $\texttt{Tile}^{M}(\cdot)$ denotes $M$ copies of the feature vector.

$\mathcal{F}_{\bm{D}}$ is used to deform the deep prior $\mathcal{F}_{\mathcal{Q}_{c}}$ to match $\mathcal{V}$ in the feature space. Thereby, according to (\ref{Eqn:DeepShapeDeformation}), we have $\mathcal{F}_{\mathcal{Q}_{v}} = \mathcal{F}_{\mathcal{Q}_{c}} + \mathcal{F}_{\bm{D}}$, with a global feature $\bm{f}_{\mathcal{Q}_{v}}$ generated by averaging $\mathcal{F}_{\mathcal{Q}_{v}}$ over $M$ points. Then $\bm{A}$ could be learned from the fusion of $\mathcal{F}_{\mathcal{I}_{o}}$, $\mathcal{F}_{\mathcal{P}_{o}}$, and $N$ copies of $\bm{f}_{\mathcal{Q}_{v}}$, via another MLP as follows:
\begin{eqnarray}
    \begin{matrix}
        \bm{A} = \texttt{MLP}([\mathcal{F}_{\mathcal{I}_{o}}, \mathcal{F}_{\mathcal{P}_{o}}, \texttt{Tile}^{N}(\bm{f}_{\mathcal{Q}_{v}})]),\\
        s.t. \quad \bm{f}_{\mathcal{Q}_{v}} = \texttt{AvgPool}(\mathcal{F}_{\mathcal{Q}_{v}}) = \texttt{AvgPool}(\mathcal{F}_{\mathcal{Q}_{c}} + \mathcal{F}_{\bm{D}}).
    \end{matrix}
\end{eqnarray}
Compared to the common practice in \cite{SPD,CRNet,SGPA} to learn $\bm{A}$ by fusing $\mathcal{F}_{\mathcal{I}_{o}}$ and $\mathcal{F}_{\mathcal{P}_{o}}$ with $N$ copies of $\bm{f}_{\mathcal{Q}_{c}}=\texttt{AvgPool}(\mathcal{F}_{\mathcal{Q}_{c}})$, our deformed version of the deep prior $\bm{F}_{\mathcal{Q}_{c}}$ via adding $\mathcal{F}_{\bm{D}}$ could effectively improve the quality of $\bm{A}$.

We also learn $\mathcal{Q}_{v}$ in (\ref{Eqn:ShallowShapeDeformation}) from $\mathcal{F}_{\mathcal{Q}_{v}}$ as follows:
\begin{eqnarray}
    \mathcal{Q}_{v} = \texttt{MLP}(\mathcal{F}_{\mathcal{Q}_{v}}) = \texttt{MLP}(\mathcal{F}_{\mathcal{Q}_{c}}+\mathcal{F}_{\bm{D}}),
\end{eqnarray}
such that according to (\ref{Eqn:ShallowShapeDeformation}) and (\ref{Eqn:DeepShapeDeformation}), we have $\mathcal{Q}_{o} = \bm{A} \times \mathcal{Q}_{v}$ and $\mathcal{F}_{\mathcal{Q}_{o}} = \bm{A} \times \mathcal{F}_{\mathcal{Q}_{v}}$, respectively. Supervisions on $\mathcal{Q}_{v}$ and $\mathcal{Q}_o$ could guide the learning of $\mathcal{F}_{\bm{D}}$ and $\bm{A}$.


\subsubsection{Pose and Size Estimator} Through the module of Deep Prior Deformer, we establish point-to-point correspondence for the observed $\mathcal{P}_{o}$  with $\mathcal{F}_{\mathcal{P}_{o}}$ by learning $\mathcal{Q}_{o}$ in the canonical space with $\mathcal{F}_{\mathcal{Q}_{o}}$. As shown in Fig. \ref{fig:DPDN}, for estimating object pose and size, we firstly pair the correspondence via feature concatenation and apply an MLP to lift the features as follows:
\begin{eqnarray}
    \mathcal{F}_{corr} = \texttt{MLP}([\mathcal{F}_{\mathcal{I}_{o}}, \texttt{MLP}(\mathcal{P}_o), \mathcal{F}_{\mathcal{P}_{o}}, \texttt{MLP}(\mathcal{Q}_o), \mathcal{F}_{\mathcal{Q}_{o}}]).
\end{eqnarray}
We then inject global information into the point-wise correspondence in $\mathcal{F}_{corr}$ by concatenating its averaged feature $\bm{f}_{corr}$, followed by an MLP to strengthen the correspondence; a pose-sensitive feature vector $\bm{f}_{pose}$ is learned from all the correspondence information via an average-pooling operation:
\begin{eqnarray}
\begin{matrix}
    \bm{f}_{pose} = \texttt{AvgPool}(\texttt{MLP}([\mathcal{F}_{corr}, \texttt{Tile}^{N}(\bm{f}_{corr})])), \\
    s.t. \quad \bm{f}_{corr} = \texttt{AvgPool}(\mathcal{F}_{corr}).
\end{matrix}
\end{eqnarray}
Finally, we apply three parallel MLPs to regress $\bm{R}$, $\bm{t}$, and $\bm{s}$, respectively:
\begin{eqnarray}
    \bm{R}, \bm{t}, \bm{s} = \rho(\texttt{MLP}(\bm{f}_{pose})), \texttt{MLP}(\bm{f}_{pose}), \texttt{MLP}(\bm{f}_{pose}),
\label{Eqn:pose-estimator}
\end{eqnarray}
where we choose a 6D representation of rotation \cite{zhou2019continuity} as the regression target of the first MLP, for its continuous learning space in $SO(3)$, and $\rho(\cdot)$ represents transformation from the 6D representation to the $3\times 3$ rotation matrix $\bm{R}$.

For the whole DPDN, we could summarize it as follows:
\begin{equation}
    \bm{R}, \bm{t}, \bm{s}, \mathcal{Q}_{v}, \mathcal{Q}_{o} = \mathrm\Phi(\mathcal{I}_{o}, \mathcal{P}_{o}, \mathcal{Q}_{c}).
\end{equation}


\subsection{Self-Supervised Training Objective $\mathcal{L}_{self-supervised}$}
\label{subsec: DPDN_self_supervised}

For an observed point set $\mathcal{P}_{o} = \{\bm{p}_{o}^{(j)}\}_{j=1}^N$, if we transform it with ($\Delta \bm{R}_1$, $\Delta\bm{t}_1$, $\Delta s_1$) and ($\Delta \bm{R}_2$, $\Delta\bm{t}_2$, $\Delta s_2$), we can obtain $\mathcal{P}_{o,1} = \{\bm{p}_{o,1}^{(j)}\}_{j=1}^N = \{\frac{1}{\Delta s_1}\Delta\bm{R}_1^T(\bm{p}_{o}^{(j)}-\Delta \bm{t}_1)\}_{j=1}^N$ and $\mathcal{P}_{o,2} = \{\bm{p}_{o,2}^{(j)}\}_{j=1}^N = \{\frac{1}{\Delta s_2}\Delta\bm{R}_2^T(\bm{p}_{o}^{(j)}-\Delta \bm{t}_2)\}_{j=1}^N$, respectively. When inputting them into DPDN in parallel, we have
\begin{align}
    &\bm{R}_{\mathcal{P}_{o,1}}, \bm{t}_{\mathcal{P}_{o,1}}, \bm{s}_{\mathcal{P}_{o,1}}, \mathcal{Q}_{v,1}, \mathcal{Q}_{o,1} = \mathrm\Phi(\mathcal{I}_{o}, \mathcal{P}_{o,1}, \mathcal{Q}_{c}), \label{Eqn:Delta1DPDN}\\
    \text{and} \quad &\bm{R}_{\mathcal{P}_{o,2}}, \bm{t}_{\mathcal{P}_{o,2}}, \bm{s}_{\mathcal{P}_{o,2}}, \mathcal{Q}_{v,2}, \mathcal{Q}_{o,2} = \mathrm\Phi(\mathcal{I}_{o}, \mathcal{P}_{o,2}, \mathcal{Q}_{c}), \label{Eqn:Delta2DPDN}
\end{align}
with $\mathcal{Q}_{v,1}=\{\bm{q}_{v,1}^{(j)}\}_{j=1}^M$, $\mathcal{Q}_{o,1}=\{\bm{q}_{o,1}^{(j)}\}_{j=1}^N$, $\mathcal{Q}_{v,2}=\{\bm{q}_{v,2}^{(j)}\}_{j=1}^M$, and $\mathcal{Q}_{o,2}=\{\bm{q}_{o,2}^{(j)}\}_{j=1}^N$.
    
There exist two solutions to $(\bm{R}, \bm{t}, \bm{s})$ of $\mathcal{P}_{o}$ from (\ref{Eqn:Delta1DPDN}) and (\ref{Eqn:Delta2DPDN}), respectively; for clarity, we use subscripts `1' and `2' for $(\bm{R}, \bm{t}, \bm{s})$ to distinguish them:

1) $\bm{R}_1, \bm{t}_1, \bm{s}_1 = \Delta \bm{R}_1 \bm{R}_{\mathcal{P}_{o,1}}, \Delta \bm{t}_1+\Delta s_1 \Delta \bm{R}_{1} \bm{t}_{\mathcal{P}_{o,1}}, \Delta s_1 \bm{s}_{\mathcal{P}_{o,1}}$;

2) $\bm{R}_2, \bm{t}_2, \bm{s}_2 = \Delta \bm{R}_2 \bm{R}_{\mathcal{P}_{o,2}}, \Delta \bm{t}_2+\Delta s_2 \Delta \bm{R}_{2} \bm{t}_{\mathcal{P}_{o,2}}, \Delta s_2 \bm{s}_{\mathcal{P}_{o,2}}$.

\noindent Upon the above parallel learning of (\ref{Eqn:Delta1DPDN}) and (\ref{Eqn:Delta2DPDN}), we design a novel self-supervised objective for the unlabeled real-world data to reduce the Sim2Real domain gap. Specifically, it combines an inter-consistency term $\mathcal{L}_{inter}$ with two intra-consistency ones $(\mathcal{L}_{intra,1}, \mathcal{L}_{intra,2})$ as follows: 
\begin{equation}
    \mathcal{L}_{self-supervised} = \lambda_1 \mathcal{L}_{inter} + \lambda_2 (\mathcal{L}_{intra,1}+\mathcal{L}_{intra,2}),
\end{equation}
where $\lambda_1$ and $\lambda_2$ are superparameters to balance the loss terms. $\mathcal{L}_{inter}$ enforces consistency across the parallel learning from $\mathcal{P}_o$ with different transformations, making the learning aware of pose changes to improve the precision of predictions, while $\mathcal{L}_{intra,1}$ and $\mathcal{L}_{intra,2}$ enforce self-adaptation between correspondence and predictions within each learning, respectively, in order to realize more reliable predictions inferred from the correspondence.

\subsubsection{Inter-Consistency Term} 

We construct the inter-consistency loss based on the following two facts: 1) two solutions to the pose and size of $\mathcal{P}_{o}$ from those of $\mathcal{P}_{o,1}$ and $\mathcal{P}_{o,2}$ are required to be consistent; 2) as representations of a same object $\mathcal{V}$ in the canonical space, $\mathcal{Q}_{v,1}$ and $\mathcal{Q}_{v,2}$ should be invariant to any pose transformations, and thus keep consistent to each other, as well as $\mathcal{Q}_{o,1}$ and $\mathcal{Q}_{o,2}$. Therefore, with two input transformations, the inter-consistency loss $\mathcal{L}_{inter}$ could be formulated as follows: 
\begin{equation}
    \mathcal{L}_{inter} = \mathcal{D}_{pose}(\bm{R}_1, \bm{t}_1, \bm{s}_1, \bm{R}_2, \bm{t}_2, \bm{s}_2) + \beta_1 \mathcal{D}_{cham}(\mathcal{Q}_{v,1}, \mathcal{Q}_{v,2}) + \beta_2 \mathcal{D}_{L2}(\mathcal{Q}_{o,1}, \mathcal{Q}_{o,2}),
\label{Eqn:consistent-loss}
\end{equation}
where $\lambda_1$ and $\lambda_2$ are balanced parameters, and
\begin{align}
    &\mathcal{D}_{pose}(\bm{R}_1, \bm{t}_1, \bm{s}_1, \bm{R}_2, \bm{t}_2, \bm{s}_2)  =  ||\bm{R}_1-\bm{R}_2||_{2} + ||\bm{t}_1-\bm{t}_2||_{2} + ||\bm{s}_1-\bm{s}_2||_{2}, \notag \\ 
    &\mathcal{D}_{cham}(\mathcal{Q}_{v,1}, \mathcal{Q}_{v,2})  = \frac{1}{2M} (\sum_{j=1}^M \min_{\bm{q}_{v,2}} ||\bm{q}_{v,1}^{(j)} - \bm{q}_{v,2}||_2 + \sum_{j=1}^M \min_{\bm{q}_{v,1}} ||\bm{q}_{v,1} - \bm{q}_{v,2}^{(j)}||_2), \notag\\
    &\mathcal{D}_{L2}(\mathcal{Q}_{o,1}, \mathcal{Q}_{o,2}) =  \frac{1}{N} \sum_{j=1}^N || \bm{q}_{o,1}^{(j)} - \bm{q}_{o,2}^{(j)} ||_{2}. \notag
\end{align}
Chamfer distance $\mathcal{D}_{cham}$ is used to restrain the distance of two complete point sets $\mathcal{Q}_{v,1}$ and $\mathcal{Q}_{v,2}$, while for the partial $\mathcal{Q}_{o,1}$ and $\mathcal{Q}_{o,2}$, we use a more strict metric of L2 distance $\mathcal{D}_{L2}$ for point-to-point constraints, since their points should be ordered to correspond with those of $\mathcal{P}_{o}$.

\subsubsection{Intra-Consistency Terms} 

For an observation $(\mathcal{I}_{o}, \mathcal{P}_{o})$, DPDN learns deep correspondence between $\mathcal{P}_{o}=\{\bm{p}_{o}^{(j)}\}_{j=1}^N$ and $\mathcal{Q}_{o}=\{\bm{q}_{o}^{(j)}\}_{j=1}^N$ to predict their relative pose and size $(\bm{R}, \bm{t}, \bm{s})$; ideally, for $\forall j=1,\cdots,N$, $\bm{q}_{o}^{(j)}=\frac{1}{||\bm{s}||_2}\bm{R}^T(\bm{p}_{o}^{(j)}-\bm{t})$. Accordingly, the predictions $\mathcal{Q}_{o,1}$ and $\mathcal{Q}_{o,2}$ in (\ref{Eqn:Delta1DPDN}) and (\ref{Eqn:Delta2DPDN}) should be restrained to be consistent with $\mathcal{Q}_{o,1}^{\prime} = \{\frac{1}{||\bm{s}_1||_2}\bm{R}_1^T(\bm{p}_{o}^{(j)}-\bm{t}_1) \}_{j=1}^N $ and $\mathcal{Q}_{o,2}^{\prime} = \{\frac{1}{||\bm{s}_2||_2}\bm{R}_2^T(\bm{p}_{o}^{(j)}-\bm{t}_2) \}_{j=1}^N$, respectively, which gives the formulations of two intra-consistency terms based on Smooth-L1 distance as follows:
\begin{equation} 
    \mathcal{L}_{intra,1} = \mathcal{D}_{SL1}(\mathcal{Q}_{o,1},  \mathcal{Q}_{o,1}^{\prime}), \quad \mathcal{L}_{intra,2} = \mathcal{D}_{SL1}(\mathcal{Q}_{o,2}, \mathcal{Q}_{o,2}^{\prime}),
\end{equation}
where 
\begin{eqnarray}
    \mathcal{D}_{SL1}(\mathcal{Q}_{1}, \mathcal{Q}_{2}) = 
    \frac{1}{N} \sum_{j=1}^N \sum_{k=1}^3
    \left\{\begin{matrix}
       5(q_{1}^{(jk)} - q_{2}^{(jk)})^2, & \text{if} |q_{1}^{(jk)} - q_{2}^{(jk)}|\le 0.1  \\
       |q_{1}^{(jk)} - q_{2}^{(jk)}|-0.05, & \text{otherwise}
      \end{matrix}\right., \notag
\end{eqnarray}
with $\mathcal{Q}_{1}=\{(q_{1}^{(j1)}, q_{1}^{(j2)}, q_{1}^{(j3)})\}_{j=1}^N $ and $\mathcal{Q}_{2}=\{(q_{2}^{(j1)}, q_{2}^{(j2)}, q_{2}^{(j3)})\}_{j=1}^N $.

\subsection{Supervised Training Objective $\mathcal{L}_{supervised}$}
\label{subsec: DPDN_supervised} Given a triplet of inputs $(\mathcal{I}_{o}, \mathcal{P}_{o}, \mathcal{Q}_{c})$ along with the annotated ground truths $(\hat{\bm{R}}, \hat{\bm{t}}, \hat{\bm{s}}, \hat{\mathcal{Q}_{v}}, \hat{\mathcal{Q}_{o}})$, we generate dual input triplets by applying two rigid transformations to $\mathcal{P}_{o}$, as done in Sec. \ref{subsec: DPDN_self_supervised}, and use the following supervised objective on top of the parallel learning of (\ref{Eqn:Delta1DPDN}) and (\ref{Eqn:Delta2DPDN}):
\begin{align}
    \mathcal{L}_{supervised}  = & \mathcal{D}_{pose}(\bm{R}_1, \bm{t}_1, \bm{s}_1, \hat{\bm{R}}, \hat{\bm{t}}, \hat{\bm{s}}) + \mathcal{D}_{pose}(\bm{R}_2, \bm{t}_2, \bm{s}_2, \hat{\bm{R}}, \hat{\bm{t}}, \hat{\bm{s}}) \notag\\
     & +  \gamma_{1}(\mathcal{D}_{cham} (\mathcal{Q}_{v,1}, \hat{\mathcal{Q}_{v}}) + \mathcal{D}_{cham} (\mathcal{Q}_{v,2}, \hat{\mathcal{Q}_{v}})) \notag\\
     & +  \gamma_{2}(\mathcal{D}_{SL1}(\mathcal{Q}_{o,1}, \hat{\mathcal{Q}_{o}}) + \mathcal{D}_{SL1}(\mathcal{Q}_{o,2}, \hat{\mathcal{Q}_{o}})).
\label{Eqn:supervised_objective}
\end{align}
We note that this supervision also implies inter-consistency between the parallel learning defined in (\ref{Eqn:consistent-loss}), making DPDN more sensitive to pose changes.

\section{Experiments}

\subsubsection{Datasets} We train DPDN on both training sets of synthetic CAMERA25 and real-world REAL275 datasets \cite{NOCS}, and conduct evaluation on REAL275 test set. CAMERA25 is created by a context-aware mixed reality approach, which renders $1,085$ synthetic object CAD models of 6 categories to real-world backgrounds, yielding a total of $300,000$ RGB-D images, with $25,000$ ones of $184$ objects set aside for validation. REAL275 is a more challenging real-world dataset, which includes $4,300$ training images of $7$ scenes and $2,754$ testing ones of $6$ scenes. Both datasets share the same categories, yet impose large domain gap.

\subsubsection{Implementation Details} To obtain instance masks for both training and test sets of REAL275, we train a MaskRCNN \cite{MasRCNN} with a backbone of ResNet101 \cite{ResNet} on CAMERA25; for settings of available training mask labels, we use the same segmentation results as \cite{SPD,DualPoseNet,SS-Conv} to make fair comparisons. We employ the shape priors released by \cite{SPD}. 

For DPDN, we resize the image crops of object observations as $192\times 192$, and set the point numbers of shape priors and observed point sets as $M=N=1,024$, respectively. In Triplet Feature Extractor, a PSP Network \cite{PSPNet} based on ResNet-18 \cite{ResNet} and two networks of PointNet++ \cite{pointnet++} are employed, sharing the same architectures as those in \cite{PVN3D,SGPA}. To aggregate multi-scale features, each PointNet++ is built by stacking $4$ set abstract levels with multi-scale grouping. The output channels of point-wise features $(\mathcal{F}_{\mathcal{I}_{o}}, \mathcal{F}_{\mathcal{P}_{o}}, \mathcal{F}_{\mathcal{Q}_{c}})$ are all set as $d=128$; other network specifics are also given in Fig. \ref{fig:DPDN}. We use ADAM to train DPDN with a total of $120,000$ iterations; the data size of a mini training batch is $B=24$ with $B_1:B_2=3:1$. The superparameters of $\lambda_1$, $\lambda_2$, $\beta_1$, $\beta_2$, $\gamma_1$ and $\gamma_2$ are set as $0.2$, $0.02$, $5.0$, $1.0$, $5.0$ and $1.0$, respectively. For each pose transformation, $\Delta \bm{R}$ is sampled from the whole SO(3) space by randomly generating three euler angles, while $\Delta \bm{t} \in \mathbb{R}^3$ with each element $\Delta t \sim U(-0.02, 0.02)$, and $\Delta s \sim U(0.8, 1.2)$.


\subsubsection{Evaluation Metrics} Following \cite{NOCS}, we report mean Average Precision (mAP) of Intersection over Union (IoU) for object detection, and mAP of $n\degree m$ cm for 6D pose estimation. IoU$_x$ denotes precision of predictions with IoU over a threshold of $x\%$, and $n\degree m$ cm denotes precision of those with rotation error less than $n\degree$ and transformation error less than $m$ cm.

\subsection{Comparisons with Existing Methods}
\label{subsec:sota}

\begin{table}[t]
    \begin{center}
    \caption{Quantitative comparisons of different methods for category-level 6D pose and size estimation on REAL275 \cite{NOCS}.  `Syn' and `Real' denote the uses of training data of synthetic CAMERA25 and real-world REAL275 datasets, respectively. `$*$' denotes training with mask labels.}
    \label{table:real275-sota}
    \resizebox{1.0\textwidth}{!}{
    \begin{tabular}{c|c|c|c|cccccc}
    \toprule
    \multirow{2}{*}{Method}& \multirow{2}{*}{Syn} & Real & Real &\multicolumn{6}{c}{mAP}\\
    \cline{5-10}
    & & w/o Label & with Label & IoU$_{50}$ & IoU$_{75}$ & 5\degree 2cm & 5\degree 5cm & 10\degree 2cm & 10\degree 5cm\\
    \midrule
    \multicolumn{10}{c}{\textbf{Unsupervised}}\\
    \midrule
    NOCS \cite{NOCS} & \checkmark &  & & 36.7 & 3.4 & - & 3.4 & - & 20.4 \\
    SPD \cite{NOCS}& \checkmark &  & & 71.0 & 
    43.1 & 11.4 & 12.0 & 33.5 & 37.8 \\
    DualPoseNet \cite{DualPoseNet} & \checkmark  & & & 68.4 & 49.5 & 15.9 & 27.1 & 33.1 & 56.8\\
    DPDN (Ours) & \checkmark &  & & 71.7 & 60.8 & 29.7 & 37.3 & 53.7 & 67.0\\
    Self-DPDN (Ours) & \checkmark  & \checkmark & & $\mathbf{72.6}$ & $\mathbf{63.8}$ & $\mathbf{37.8}$ & $\mathbf{45.5}$ & $\mathbf{59.8}$ & $\mathbf{71.3}$\\
    \hline
    UDA-COPE \cite{UDA-COPE}& \checkmark  & \checkmark$^*$ & & 82.6 & 62.5 & 30.4 & 34.8 & 56.9 & 66.0 \\
    Self-DPDN (Ours) & \checkmark  & \checkmark$^*$ & & $\mathbf{83.0}$ & $\mathbf{70.3}$ & $\mathbf{39.4}$ & $\mathbf{45.0}$ & $\mathbf{63.2}$ & $\mathbf{72.1}$  \\
    \midrule
    \multicolumn{10}{c}{\textbf{Supervised}}\\
    \midrule
    NOCS \cite{NOCS} & \checkmark & & \checkmark  & 78.0 & 30.1 & 7.2 & 10.0 & 13.8 & 25.2 \\
    SPD  \cite{SPD} & \checkmark & & \checkmark & 77.3 & 53.2 & 19.3 & 21.4 & 43.2 & 54.1 \\
    CR-Net \cite{CRNet} & \checkmark & & \checkmark & 79.3 & 55.9 & 27.8 & 34.3 & 47.2 & 60.8 \\
    DualPoseNet \cite{DualPoseNet} & \checkmark &  & \checkmark & 79.8 & 62.2 & 29.3 & 35.9 & 50.0 & 66.8 \\
    SAR-Net \cite{lin2022sar}& \checkmark & & \checkmark & 79.3 & 62.4 & 31.6 & 42.3 & 50.3 & 68.3 \\
    SS-ConvNet \cite{SS-Conv}& \checkmark & & \checkmark & 79.8 & 65.6 & 36.6 & 43.4  & 52.6 & 63.5 \\
    SGPA \cite{SGPA}         & \checkmark & & \checkmark & 80.1 & 61.9 & 35.9 & 39.6 & 61.3 & 70.7 \\
    DPDN (Ours) & \checkmark & & \checkmark & $\mathbf{83.4}$ & $\mathbf{76.0}$ & $\mathbf{46.0}$ & $\mathbf{50.7}$ & $\mathbf{70.4}$ & $\mathbf{78.4}$\\
    \bottomrule
    \end{tabular}}
    \end{center}
\end{table}

We compare our method with the existing ones for category-level 6D object pose and size estimation under both unsupervised and supervised settings. Quantitative results are given in Table \ref{table:real275-sota}, where results under supervised setting significantly benefit from the annotations of real-world data, compared to those under unsupervised one; for example, on the metric of $5\degree 2$ cm, SPD \cite{SPD} improves the results from $11.4\%$ to $19.3\%$, while DualPoseNet \cite{DualPoseNet} improves from $15.9\%$ to $29.3\%$. Therefore, the exploration of UDA for the target task in this paper is of great practical significance, due to the difficulties in precisely annotating real-world object instances in 3D space.

\subsubsection{Unsupervised Setting}

Firstly, a basic version of DPDN is trained on the synthetic data and transferred to real-world domain for evaluation; under this setting, our basic DPDN outperforms the existing methods on all the evaluation metrics, as shown in Table \ref{table:real275-sota}. To reduce the Sim2Real domain gap, we further include the unlabeled Real275 training set via our self-supervised DPDN for UDA; results in Table \ref{table:real275-sota} verify the effectiveness of our self-supervised DPDN (dubbed \textbf{Self-DPDN} in the table), which significantly improves the precision of the basic version, \textit{e.g.}, a performance gain of $8.1\%$ on $5\degree 2$ cm from $29.7\%$ to $37.8\%$.

UDA-COPE \cite{UDA-COPE} is the first work to introduce the unsupervised setting, which trains deep model with a teacher-student scheme to yield pseudo labels for real-world data; in the process of training, pose annotations of real-world data are not employed, yet mask labels are used for learning instance segmentation. To fairly compare with UDA-COPE, we evaluate DPDN under the same setting; results in Table \ref{table:real275-sota} also show the superiority of our self-supervised DPDN over UDA-COPE, especially for the metrics of high precisions, \textit{e.g.}, an improvement of $10.2\%$ on $5\degree 5$ cm. The reason for the great improvement is that UDA-COPE heavily relies on the qualities of pseudo labels, while our self-supervised objective could guide the optimization 
moving for the direction meeting inter-/intra- consistency, to make the learning fit the characteristics of REAL275 and decrease the downside effect of the synthetic domain.

\subsubsection{Supervised Setting}

We also compare our DPDN with the existing methods, including those of direct regression \cite{DualPoseNet,SS-Conv}, and those based on dense correspondence learning\cite{NOCS,SPD,CRNet,lin2022sar,SGPA}, under supervised setting. As shown in Table \ref{table:real275-sota}, DPDN outperforms the existing methods on all the evaluation metrics, \textit{e.g.}, reaching the precisions of $76.0\%$ on IoU$_{75}$ and $78.4\%$ on $10\degree 5$ cm. Compared with the representative SS-ConvNet \cite{SS-Conv}, which directly regresses object poses and sizes from observations, our DPDN takes the advantages of categorical shape priors, and achieves more precise results by regressing from deep correspondence; compared with SGPA \cite{SGPA}, the recent state-of-the-art method based on correspondence learning (\textit{c.f.} Eq. (\ref{Eqn:ShallowShapeDeformation})), our DPDN shares the same feature extractor, yet benefits from the direct objectives for pose and size estimation, rather than the surrogate ones, \textit{e.g.}, for regression of $\bm{D}$ and $\bm{A}$ in Eq. (\ref{Eqn:ShallowShapeDeformation}); DPDN thus achieves more reliable predictions.

\begin{table}[t]
    \begin{center}
    \caption{Ablation studies on the variants of our proposed DPDN under supervised setting. Experiments are evaluated on REAL275 test set \cite{NOCS}.}
    \label{table:supervised-ablation}
    \begin{tabular}{>{\centering}p{1cm}|>{\centering}p{1cm}|c|c|cccccc}
    \toprule
    \multicolumn{2}{c|}{Input} & Deep Prior & Pose$ \& $Size  & \multicolumn{6}{c}{mAP}\\
    \cline{1-2} \cline{5-10}
    $\mathcal{P}_{o,1}$ & $\mathcal{P}_{o,2}$ & Deformer & Estimator& IoU$_{50}$ & IoU$_{75}$ & 5\degree 2cm & 5\degree 5cm & 10\degree 2cm & 10\degree 5cm\\
    \midrule
    \checkmark & $\times$ &  $\times$     & \checkmark & 79.9 & 65.4 & 26.7 & 35.6 & 47.2 & 63.9 \\
    \checkmark & \checkmark &  $\times$   & \checkmark &83.3 & 72.9 & 35.2 & 43.9 & 57.3 & 70.8 \\
    \checkmark & $\times$ &  \checkmark   & \checkmark &$\mathbf{83.4}$ & $\mathbf{76.2}$ & 39.6 & 46.1 & 65.5 & 76.7 \\
    \checkmark & \checkmark &  \checkmark & \checkmark &$\mathbf{83.4}$ & 76.0 & $\mathbf{46.0}$ & $\mathbf{50.7}$ & $\mathbf{70.4}$ & $\mathbf{78.4}$ \\
    \hline
    \checkmark & \checkmark &  \checkmark &  $\times$ &  59.6 & 45.6 & 27.9 & 33.0 & 50.2 & 63.9 \\
    \bottomrule
    \end{tabular}
    \end{center}
\end{table}

\subsection{Ablation Studies and Analyses}

In this section, we conduct experiments to evaluate the efficacy of both the designs in our DPDN and the self-supervision upon DPDN. 

\subsubsection{Effects of the Designs in DPDN} We verify the efficacy of the designs in DPDN under supervised setting, with the results of different variants of our DPDN shown in Table. \ref{table:supervised-ablation}. Firstly, we confirm the effectiveness of our Deep Prior Deformer with categorical shape priors; by removing Deep Prior Deformer, the precision of DPDN with parallel learning on $5\degree 2$ cm drops from $46.0\%$ to $35.2\%$, indicating that learning from deep correspondence by deforming priors in feature space indeed makes the task easier than that directly from object observations.

Secondly, we show the advantages of using true objectives for direct estimates of object poses and sizes, over the surrogate ones for the learning of the canonical point set $\mathcal{Q}_o$ to pair with the observed $\mathcal{P}_{o}$. Specifically, we remove our Pose and Size Estimator, and make predictions by solving Umeyama algorithm to align $\mathcal{P}_{o}$ and $\mathcal{Q}_o$; precisions shown in the table decline sharply on all the evaluation metrics, especially on IoU$_x$. We found that the results on IoU$_x$ are also much lower than those methods based on dense correspondence learning \cite{SPD,CRNet,SGPA}, while results on $n\degree m$ are comparable; the reason is that we regress the absolute coordinates of $\mathcal{Q}_c$, rather than the deviations $\bm{D}$ in Eq. (\ref{Eqn:ShallowShapeDeformation}), which may introduce more outliers to affect the object size estimation.

Thirdly, we confirm the effectiveness of the parallel supervisions in (\ref{Eqn:supervised_objective}), \textit{e.g.}, inputting $\mathcal{P}_{o,1}$ and $\mathcal{P}_{o,2}$ of the same instance with different poses. As shown in Table \ref{table:supervised-ablation}, results of DPDN with parallel learning are improved (with or without Deep Prior Deformer), since the inter-consistency between dual predictions is implied in the parallel supervisions, making the learning aware of pose changes.

\begin{table}[t]
    \begin{center}
    \caption{Ablation studies of our proposed self-supervised objective upon DPDN under unsupervised setting. Experiments are evaluated on REAL275 test set \cite{NOCS}.}
    \label{table:self-supervised-ablation}
    \begin{tabular}{>{\centering}p{1.5cm}|>{\centering}p{1.5cm}|cccccc}
    \toprule
    \multirow{2}{*}{$\mathcal{L}_{inter}$} & $\mathcal{L}_{intra,1}$ & \multicolumn{6}{c}{mAP}\\
    \cline{3-8}
    & $\mathcal{L}_{intra,2}$ & IoU$_{50}$ & IoU$_{75}$ & 5\degree 2cm & 5\degree 5cm & 10\degree 2cm & 10\degree 5cm\\
    \midrule
    $\times$ & $\times$ &  71.7 & 60.8 & 29.7 & 37.3 & 53.7 & 67.0\\
    \checkmark & $\times$ &  $\mathbf{72.6}$ & 63.2 & 36.9 & 43.7 & 58.7 & 68.7 \\
    $\times$ & \checkmark & 70.9 & 58.6 & 35.8 & 43.6 & 56.6 & 69.3\\
    \checkmark & \checkmark & $\mathbf{72.6}$ & $\mathbf{63.8}$ & $\mathbf{37.8}$ & $\mathbf{45.5}$ & $\mathbf{59.8}$ & $\mathbf{71.3}$\\
    \bottomrule
    \end{tabular}
    \end{center}
\end{table}

\subsubsection{Effects of the Self-Supervision upon DPDN} 
\begin{figure}[t]
    \centering
    \includegraphics[width=0.83\linewidth]{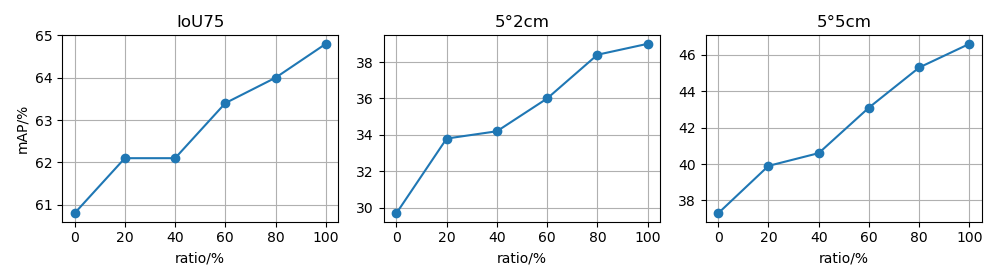}
    \caption{Plottings of mAP versus the ratio of unlabeled REAL275 training data under unsupervised setting. Experiments are evaluated on REAL275 test set \cite{NOCS}.}
    \label{fig:ratio}
\end{figure}

We have shown the superiority of our novel self-supervised DPDN under the unsupervised setting in Sec. \ref{subsec:sota}; here we include the evaluation on the effectiveness of each consistency term in the self-supervised objective, which is confirmed by the results shown in Table \ref{table:self-supervised-ablation}. Taking results on $5\degree,2$ cm as examples, DPDN with inter-consistency term $\mathcal{L}_{inter}$ improves the results of the baseline from $29.7\%$ to $36.9\%$, and DPDN with the intra-consistency ones $\mathcal{L}_{intra,1}$ and $\mathcal{L}_{intra,2}$ improves to $35.8\%$, while their combinations further refresh the results, revealing their strengths on reduction of domain gap. We also show the influence of data size of unlabeled real-world images on the precision of predictions in Fig. \ref{fig:ratio}, where precisions improve along with the increasing ratios of training data.

\begin{figure}[h]
    \centering
    \includegraphics[width=0.93\linewidth]{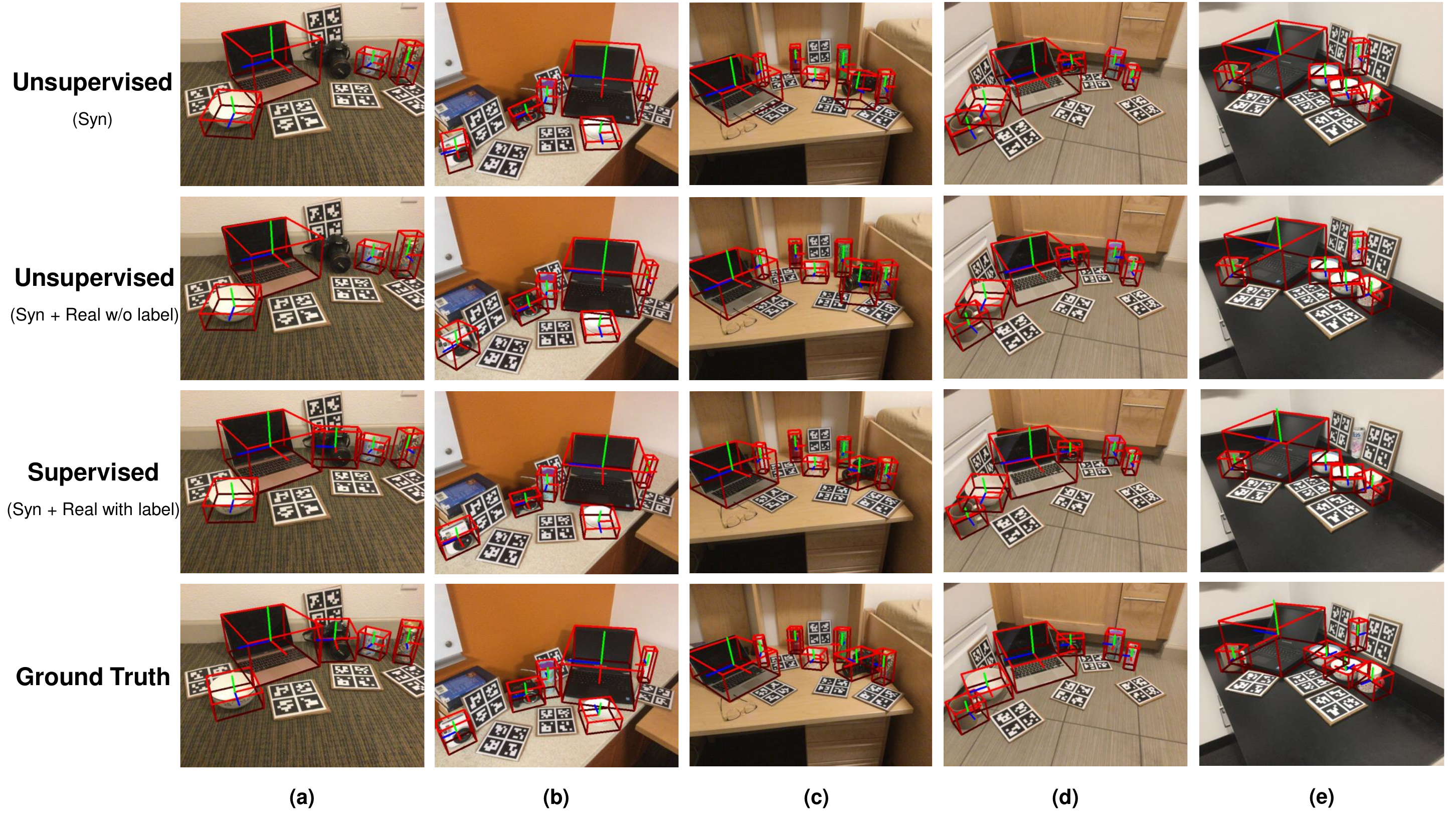}
    \caption{Qualitative results of DPDN on REAL275 test set \cite{NOCS}. `Syn' and `Real' denote the uses of synthetic CAMERA25 and real-world REAL275 training sets, respectively.}
    \label{fig:vis}
\end{figure}

\subsection{Visualization}

We visualize in Fig. \ref{fig:vis} the qualitative results of our proposed DPDN under different settings on REAL275 test set \cite{NOCS}. As shown in the figure, our self-supervised DPDN without annotations of real-world data, in general, achieves comparable results with the fully-supervised version, although there still exist some difficult examples, \textit{e.g.}, cameras in Fig. \ref{fig:vis} (a) and (b), due to the inaccurate masks from MaskRCNN trained on synthetic data. Under the unsupervised setting, our self-supervised DPDN also outperforms the basic version trained with only CAMERA25, by including unlabeled real-world data with self-supervision; for example, more precise poses of laptops are obtained in Fig. \ref{fig:vis} (c) and (d). 

~\\

\noindent\textbf{Acknowledgements} This work is supported in part by Guangdong R$\&$D key project of China (No.: 2019B010155001), and the Program for Guangdong Introducing Innovative and Enterpreneurial Teams (No.: 2017ZT07X183).

\newpage

\bibliographystyle{splncs04}
\bibliography{egbib}
\end{document}